# A Dependency Look at the Reality of Constituency

*Xinying Chen[1],*
*Carlos Gómez-Rodríguez[2],*
*Ramon Ferrer-i-Cancho[3]*

**Abstract.** A comment on "Neurophysiological dynamics of phrase-structure building during sentence processing" by Nelson et al (2017), Proceedings of the National Academy of Sciences USA 114(18), E3669-E3678.

Recently, Nelson et al. (2017) have addressed the fundamental problem of the neurophysiological support for complex syntactic operations of theoretical computational models. They interpret their compelling results as supporting the neural reality of phrase structure. Such a conclusion opens various questions.

First, constituency is not the only possible reality for the syntactic structure of sentences. An alternative is dependency, where the structure of a sentence is defined by word pairwise dependencies (Fig. 1). From that perspective, phrase structure is regarded as an epiphenomenon of word-word dependencies and constituency (in a classical sense as that of X-bar theory) has been argued to not exist (Mel'čuk, 2011). Furthermore, constituency may not be universal and thus its suitability may depend on the language (Evans & Levinson, 2009). Dependency is a stronger alternative for its simplicity, its close relationship with merge (Osborne, Putnam, & Gross, 2011), its compatibility with recent cognitive observations (Gómez-Rodríguez, 2016), its contribution to the cost of individual words even in isolation (Lester & Moscoso del Prado Martin, 2016) and its success over phrase structure in computational linguistics, where it has become predominant (Kübler, McDonald, & Nivre, 2009).

Second, the authors admit that a parser of the sentence might transiently conclude that "ten sad students"... is a phrase consistently with a transient decrease in activity (1st paragraph of p. 4). Unfortunately, their parser does not account for that as shown by the counts in Fig. 2 A of Nelson et al. (2017). In contrast, a standard dependency parser would, because at that point it would close the dependencies opened by "ten" and "sad" (Fig. 1). This raises the question of whether the conclusions depend on the choice X-bar and particular parser as a model of phrase structure. The conclusions by Nelson et al. (2017) may suffer from circularity, namely the positive support for a particular X-bar model could be due to the fact that the source was a toy artificial X-bar grammar. Future analyses would benefit from the use of natural sentences, sentences with realistic probabilities that are also longer and more complex (sentence length does not exceed 10 in Nelson et al. (2017)).

Third, dependency shows the limits of comparing phrase structure models against *n*-gram models with $n = 2$, because only about 50% of adjacent words are linked (Liu, 2008; Ferrer-i-

---

[1] Foreign Languages Research Center, School of Foreign Studies, Xi'an Jiaotong University, No.28 Xianning West Road, 710049 Xi'an, Shaanxi, P.R. China.
[2] Universidade da Coruña. FASTPARSE Lab, LyS Research Group. Departamento de Computación. Facultade de Informática, Elviña 15071 A Coruña, Spain
[3] Complexity & Quantitative Linguistics Lab, LARCA Research Group, Departament de Ciències de la Computació, Universitat Politècnica de Catalunya, Campus Nord, Edifici Omega, Jordi Girona Salgado 1-3, 08034 Barcelona, Catalonia (Spain). Corresponding author, rferrericancho@cs.upc.edu.

Cancho, 2004), thus a bigram model misses 50% of the dependencies. Bigrams are a weak baseline, as the common practice in computational linguistics is using at least smoothed trigram models, and often 5-gram models, to obtain meaningful predictions (Jozefowicz, Vinyals, Schuster, Shazeer, & Wu, 2016). A higher-order lexical *n*-gram model would strengthen the current results. The authors also employ more sophisticated *n*-gram models. One is an unbounded model based on part-of-speech categories, implying a dramatic loss of information with respect to the original words which might explain its poor performance. The other is a syntactic *n*-gram, but not enough information is provided about its definition and implementation. Regardless, since the model is obtained from a corpus derived from a toy grammar and lexicon, its probabilities are likely to be unrealistic and thus it is problematic.

In sum, dependency offers a better approach to the syntactic complexity of languages and merge. *n*-gram models of higher complexity should be the subject of future research involving realistic sentences.

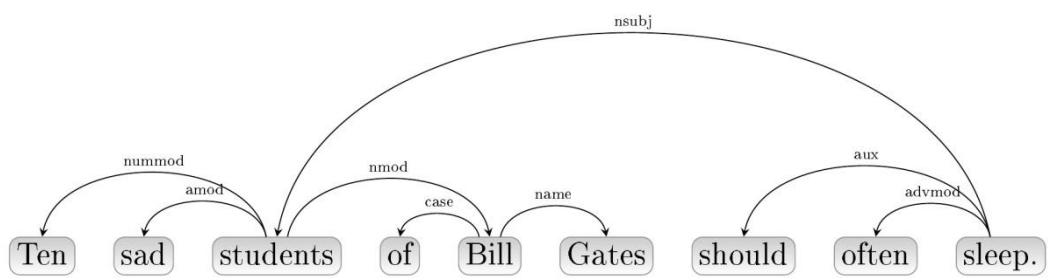

Figure 1: Syntactic dependency structure of the sentence in Fig 2 A of Nelson et al. (2017) according to Universal Dependencies (McDonald et al., 2013).

## Acknowledgements

X.C. is supported by the Social Science Fund of Shaanxi State (2015K001). C.G.R is funded by the European Research Council (ERC) under the European Union's Horizon 2020 research and innovation programme (grant agreement No 714150 FASTPARSE), and by the TELEPARES-UDC project (FFI2014-51978-C2-2-R) from MINECO (Ministerio de Economía y Competitividad). R.F.C is funded by the grants 2014SGR 890 (MACDA) from AGAUR (Generalitat de Catalunya) and the grant TIN2014-57226-P from MINECO.

## References

**Evans, N., & Levinson, S. C.** (2009). The myth of language universals: language diversity and its importance for cognitive science. *Behavioral and Brain Sciences 32, 429-492.*

**Ferrer-i-Cancho, R.** (2004). Euclidean distance between syntactically linked words. *Physical Review E 70, 056135.*

**Gómez-Rodríguez, C.** (2016). Natural language processing and the Now-or-Never bottleneck. *Behavioral and Brain Sciences 39, e74.*

**Jozefowicz, R., Vinyals, O., Schuster, M., Shazeer, N., & Wu, Y.** (2016). *Exploring the limits of language modeling.* arXiv preprint arXiv:1602.02410.

**Kübler, S., McDonald, R. & Nivre, J**. (2009). *Dependency parsing.* Morgan and Claypool Publishers.

**Lester, N. A. & Moscoso del Prado Martín, F.** (2016). Syntactic flexibility in the noun: Evidence from picture naming. In: Papafragou, A., Grodner, D., Mirman, D., & Trueswell, J.C. (Eds.), *Proceedings of the 38th Annual Conference of the Cognitive Science Society (pp. 2585-2590).* Austin, TX: Cognitive Science Society.

**Liu, H.** (2008). Dependency distance as a metric of language comprehension difficulty. *Journal of Cognitive Science 9, 159-191.*

**McDonald, R., Nivre, J., Quirmbach-Brundage, Y., Goldberg, Y., Das, D., Ganchev, K. Hall, K.B., Petrov, S., Zhang, H., Täckström, O., Bedini, C., Bertomeu, N. & Lee, J.** (2013). Universal dependency annotation for multilingual parsing. *Proceedings of ACL (pp. 92-97).*

**Mel'čuk, I**. (2011). Dependency in language-2011. In: K. Gerdes, E. Hajicova, & L. Wanner (Eds.), *Proceedings of the international conference on dependency linguistics, DepLing 2011, Barcelona, September 5-7, 2011.*

**Nelson, M. J., El Karoui, I., Giber, K., Yang, X., Cohen, L., Koopman, H., Cash, S. S., Naccache, L., Hale, J. T., Pallier, C.P. & Dehaene, S.** (2017). Neurophysiological dynamics of phrase-structure building during sentence processing. *Proceedings of the National Academy of Sciences, 114 (18), E3669-E3678.*

**Osborne, T., Putnam, M., & Gross, T.** (2011). Bare phrase structure, label-less trees, and specifier-less syntax: Is minimalism becoming a dependency grammar? *The Linguistic Review 28, 315-364.*